\documentclass[conference]{IEEEtran}
\IEEEoverridecommandlockouts
\usepackage{amsmath,amssymb,amsfonts,multirow}
\usepackage{algorithmic}
\usepackage{colortbl}
\usepackage{url}
\usepackage{cite}
\usepackage{amssymb} 
\usepackage{graphicx}
\usepackage{textcomp}
\usepackage{bbding}
\usepackage{color} 
\usepackage{xcolor}
\def\BibTeX{{\rm B\kern-.05em{\sc i\kern-.025em b}\kern-.08em
    T\kern-.1667em\lower.7ex\hbox{E}\kern-.125emX}}
\begin{document}

\title{RWKV-UI: UI Understanding with Enhanced Perception and Reasoning}

\author{Jiaxi Yang\textsuperscript{1,2} and 
  Haowen Hou\textsuperscript{\textdagger}\\
  \textsuperscript{1}Sun Yat-sen University, Shenzhen, China and \textsuperscript{2}Shenzhen Yuanshi Intelligence Co., Ltd\\
  \textsuperscript{\textdagger}Guangdong Laboratory of Artificial Intelligence and Digital Economy (SZ), Shenzhen, China \\
  \texttt{yangjx87@mail2.sysu.edu.cn}, 
  \texttt{houhaowen@gml.ac.cn}
 }

\maketitle

\begin{abstract}
Existing Visual Language Modelsoften struggle with information loss and limited reasoning abilities when handling high-resolution web interfaces that combine complex visual, textual, and interactive elements. These challenges are particularly evident in tasks requiring webpage layout comprehension and multi-step interactive reasoning. To address these challenges, we propose RWKV-UI, a Visual Language Model based on the RWKV architecture, specifically designed to handle high-resolution UI images. During model training, we introduce layout detection as a visual prompt to help the model better understand the webpage layout structures. Additionally, we design a visual prompt based on the Chain-of-Thought(CoT) mechanism, which enhances the model's ability to understand and reason about webpage content through reasoning chains. Experimental results show that RWKV-UI demonstrates significant performance improvements in high-resolution UI understanding and interactive reasoning tasks.
\end{abstract}


\begin{figure*}[ht]
    \centering
    \includegraphics[width=1\linewidth]{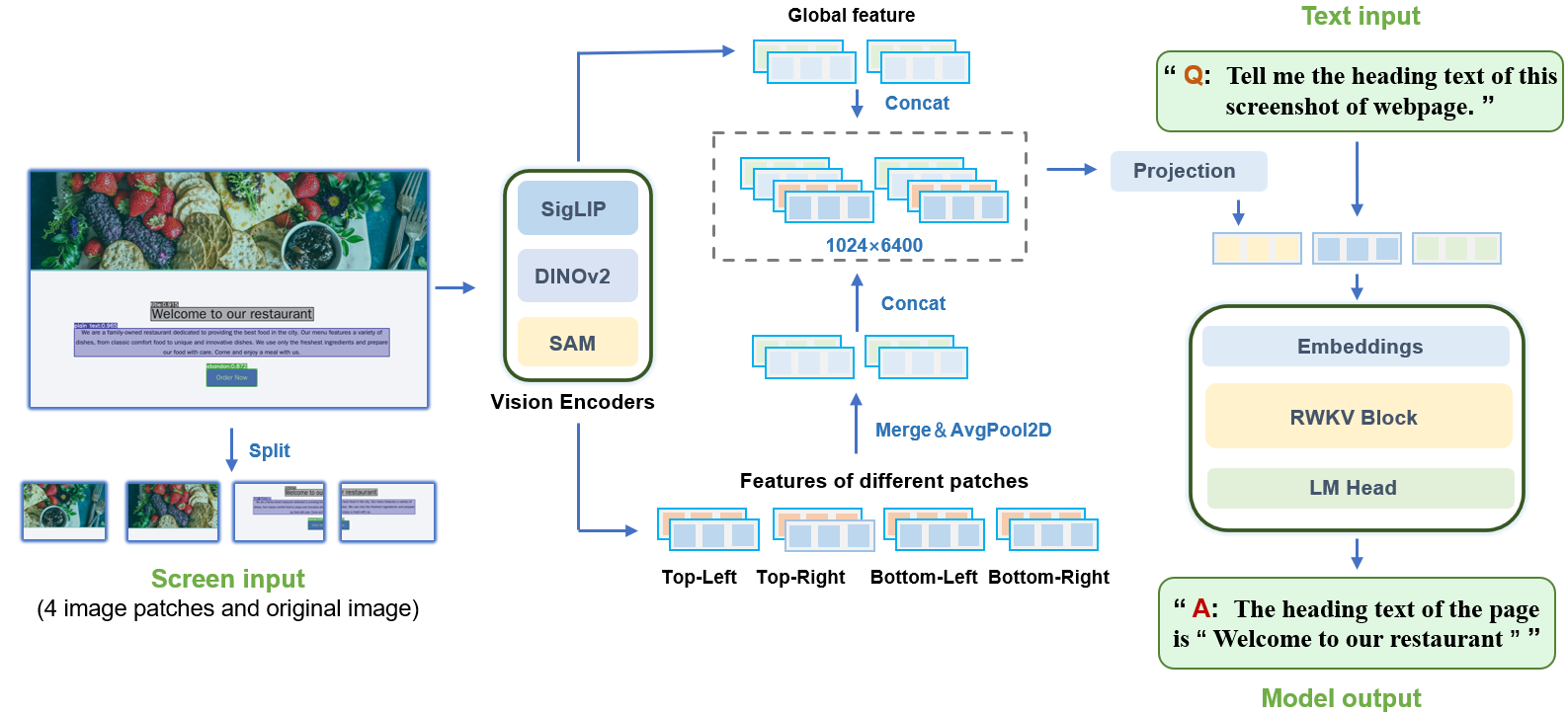}
    \caption{The overrall architecture of our model. The model input includes the complete webpage image with visual prompts, the split images of its four sections, and the textual input.}
    \label{architecture}
\end{figure*}

\section{Introduction}
\label{sec:intro}
With the rapid development of multimodal tasks, Vision Language Models(VLMs) have made significant progress in both academia and industry. Models such as LLaVA\cite{visualinstructiontuning}, DeepSeek\cite{deepseekvl}, mPLUG-Owl2\cite{mPLUG-Owl2}, Kosmos2\cite{Kosmos-2}, Bunny\cite{Bunny}, Ferret-UI\cite{Ferret-UI}, and ScreenAI\cite{ScreenAI} have demonstrated excellent capabilities in integrating visual and linguistic information, and they are widely applied in tasks such as document analysis, visual question answering, and multimodal information retrieval. Despite these successes, these models still face challenges when handling high-resolution images, preserving details, and performing interactive reasoning for UI interfaces. UI interfaces integrate visual, textual, and interactive elements such as buttons, icons, and menus, where their layout and logical relationships are crucial for precise human-computer interaction\cite{Mobile-Agent}\cite{appTesting}. However, the model struggles to recognize these elements, especially with low-resolution images, where key details are often lost. This results in blurred text, unclear boundaries, and missing element information, making interaction and accurate inference challenging.\par
To address these challenges, we proposed RWKV-UI which adopts a three-visual encoder architecture and improves reasoning efficiency by performing efficient partition encoding and feature recombination in high-resolution images. This strategy enables RWKV-UI to process UI images with resolutions up to 4096×4096 while maintaining image detail, thus enhancing the model's ability to understand fine-grained UI details.\par
Additionally, Visual Prompt Engineering introduces color and layout prompts to help the model better capture the precise relationships between images and text. For example, by marking specific color regions in images and using color-related templates in text, the model can be guided to establish clearer links between the visual and linguistic components. Building upon this, RWKV-UI also introduces layout detection and a Chain-of-Thought(CoT) reasoning mechanism. This enables RWKV-UI to understand the webpage layout information, allowing it to infer user actions and their logical relationships. This improves reasoning accuracy and interaction understanding, and subsequent experiments have shown that our strategies yield excellent results.\par
Our contributions are summarized as follows:
\begin{itemize}
  \item Visual prompt engineering: By incorporating color and layout prompts, we enable the model to establish accurate associations between visual elements and textual information, significantly improving the accuracy of UI understanding.
  \item Visually enhanced CoT reasoning mechanism: We develop a reasoning mechanism based on step-by-step question answering using visual prompts derived from website images. The process begins with identifying key elements, proceeds to inferring logical relationships, and concludes with predicting user interaction paths, thereby supporting multi-step reasoning in complex UI tasks.
  \item High-resolution model: We propose a model with three visual encoders, which combined with a high-resolution strategy, enabling the model to effectively handle high-resolution UI images.
\end{itemize}
Experimental results show that RWKV-UI outperforms existing models in several UI understanding tasks, demonstrating exceptional reasoning capabilities, especially in high-resolution scenarios. 

\section{RELATED WORK}
\subsection{Visual prompts}
In multimodal learning, prompting has become a key technique for improving VLM performance. Traditionally, text-based prompting\cite{Texts}\cite{Prompting}\cite{Conditional-Prompt} inserts learnable prompts into input text to guide models in completing tasks. Visual prompts, on the other hand, use cues like color, shape, or position to improve a model's understanding of visual information. For example, CPT\cite{CPT}  uses color prompts to assist object recognition. T-Rex\cite{T-Rex} uses visual prompts for object counting, and T-Rex2\cite{T-Rex2} encodes points or bounding boxes as embeddings to support various reasoning workflows. Similarly to CPT and T-Rex,  our work combines color-based prompts with visual-language prompts to enhance model reasoning capabilities.
\subsection{UI unstanding}
Early research on the understanding of UI focuses mainly on task execution and intelligent navigation for web interfaces\cite{Navigate-the-Web}\cite{Reinforcement-Learning}. As UI complexity increased, research shifted towards multimodal Vision-Language Models (VLMs). Ferret-UI\cite{Ferret-UI} improves visual capabilities with an 'arbitrary resolution' strategy, while ScreenAI improves performance by refining the PaLI architecture and generating large-scale datasets. Furthermore, agents based on language models, such as Mobile-Agent\cite{Mobile-Agent} and AppAgent\cite{AppAgent}, integrate visual and language information, improving reasoning and interaction in complex UI scenarios.
\subsection{Visual Language Models} 
Visual-Language Models(VLMs) combine visual and textual information to handle complex reasoning and tasks. RWKV\cite{RWKV} is an efficient RNN architecture with linear complexity and constant memory usage, achieving GPT-level performance in language modeling. VisualRWKV\cite{VisualRWKV} extends RWKV to the visual-language domain, enabling efficient joint processing of visual and textual information and demonstrating advantages in long-sequence modeling.

\section{Methodology}
\subsection{Architecture}
The architecture adopted in RWKV-UI, as shown in Fig. \ref{architecture}, now includes three visual encoders: SIGLIP, DINO, and SAM. In particular, the SAM encoder increases the maximum supported resolution to 1024×1024, significantly enhancing the model's ability to capture key features in images.
We implemented a split-encode-combine strategy: input images with visual prompts are divided into four parts, which are processed by the SIGLIP, DINOv2, and SAM encoders to extract features for each quadrant (top-left, bottom-left, top-right, and bottom-right). These features are then merged and passed through avgpool2d. The pooled features are aggregated with the global features obtained from the original (nonsplit) image processed by the encoders, resulting in a combined feature set that integrates global and fine-grained local information.

This approach enables the model to support resolutions up to 4096×4096, allowing it to efficiently and accurately handle UI understanding tasks, particularly for small and blurry elements in webpage images, such as buttons, markers, and hyperlinks. Unlike other methods that rely on token compression~\cite{mPLUG-Owl2}, our approach is completely lossless, preserving all the detailed information from the original image. Furthermore, increasing the resolution does not increase the inference cost. Whether processing 1024×1024 or 4096×4096 images, the number of image tokens used remains constant at 1024, ensuring efficient performance scaling without a trade-off in speed or accuracy. This enhancement significantly improves the model's ability to analyze and interpret complex visual information within UI interfaces, enabling more precise recognition and interaction with fine-grained details.
\begin{figure*}[ht]
    \centering
    \includegraphics[width=1\linewidth]{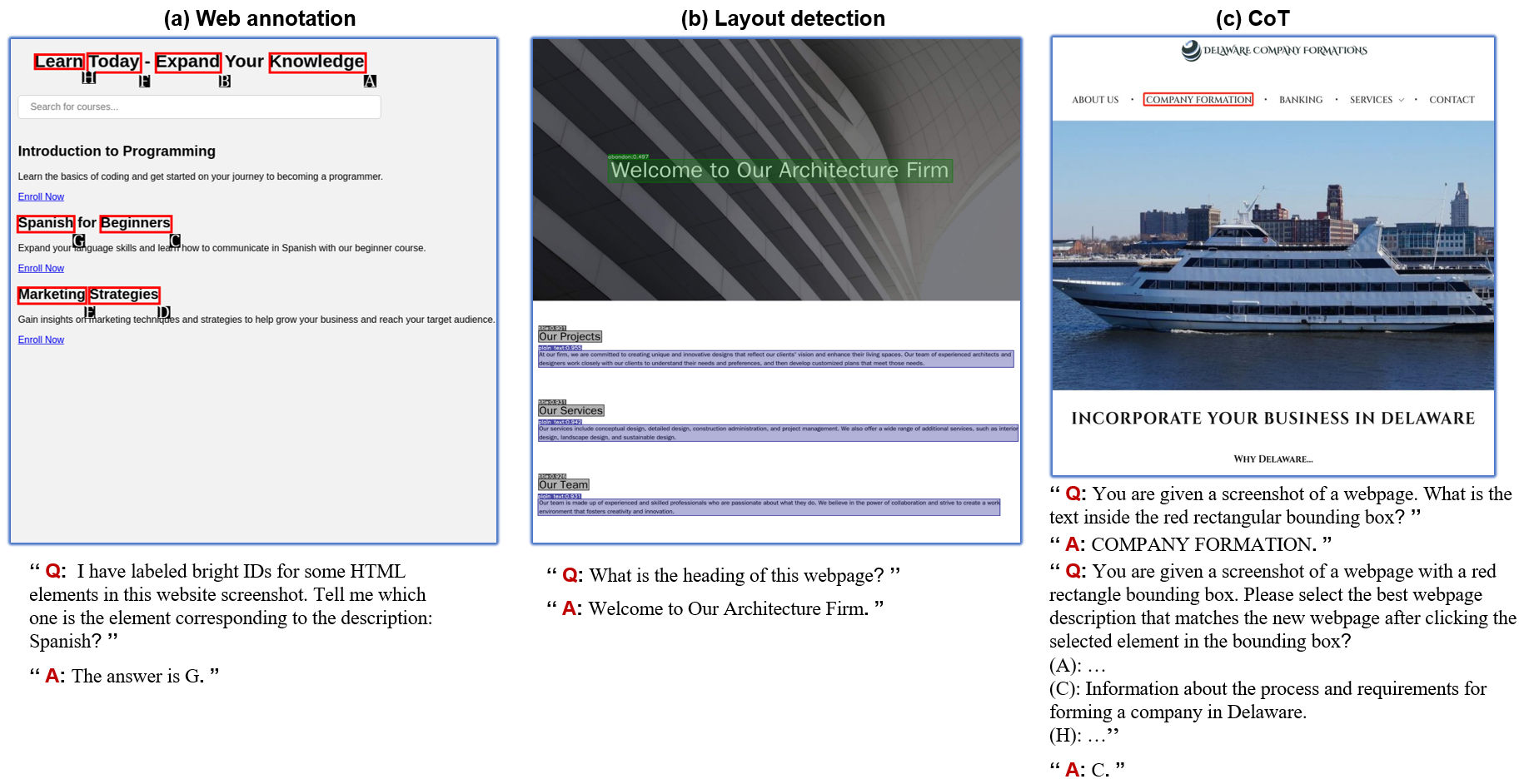}
    \caption{Data samples based on visual prompt engineering include (a) webpage annotation data, (b) layout detection-based QA data, and (c) Chain-of-Thought reasoning-based data}
    \label{data_sample}
\end{figure*}

\subsection{Data Generation}
UI interface images have more interactive elements like buttons and links than routine and document images. Just using textual prompts for interaction modeling can't accurately get the position and content of these web elements. Also, it's hard for models to reason about them, like predicting the interface after clicking a link. Such reasoning is usually beyond current models. To help models understand and reason about web interfaces better, we gathered a large dataset of webpage, PDF, and chart images. These images are carefully annotated with visual info and combined with textual prompts to boost the model's ability to interact with and understand web interfaces effectively.

\subsubsection{Visual Prompt}\label{se1}
The visual prompting framework we designed consists of two key components:
\begin{itemize}
\item Annotating and detecting of HTML elements in web pages using red bounding boxes 
\item Marking and understanding the structural layout of the web interface.
\end{itemize}
\textbf{Web Anonotation:} 
This study proposes an automated method that combines image processing and OCR technology to identify HTML elements from webpage screenshots and generate question-answer data. The process is as follows: First, the screenshot is converted to grayscale and binary format to separate the foreground from the background, and morphological operations are applied to reduce noise. Next, OpenCV's contour detection algorithm is used to locate candidate regions, and rectangles that match HTML element characteristics are filtered based on area and aspect ratio. Then, these rectangular regions are cropped from the original image, and Tesseract OCR is used to extract text content.Finally, the extracted text is denoised and cleaned to ensure accurate matching with the regions. Red rectangular boxes that meet the criteria are drawn, as shown in Fig. \ref{data_sample}(a), which displays 8 red rectangular boxes with corresponding labels. The content within these red boxes is extracted using OCR, and one box is randomly selected. The content of this selected box is used as the question, and the label of the corresponding red box is provided as the answer. This completes the process of recognizing and selecting textual content from the webpage image.\par
\textbf{Layout detection:} 
Additionally, we observed that the layout structures of PDF documents and webpage images exhibit significant similarities, as both are primarily composed of text and image content. Since complex elements such as formulas are relatively rare in webpage images, we can focus on the layout detection task without the need for in-depth analysis of specific content. Therefore, we utilized the layout detection model from the PDF-Extract-Kit\cite{mineru} to perform layout analysis on webpage images.\par
Specifically, we utilized the Doclayout-YOLO-model\cite{DocLayout-YOLO} to efficiently detect text blocks and image regions within web page images and employed these results to generate hierarchical representations of web page layouts. The detected layout elements are further labeled into different categories (such as titles, paragraphs, buttons, etc.), as shown in Fig. \ref{data_sample}(b). The text content and titles in the webpage image are highlighted with colored background boxes. By using different colors for labeling, the model's ability to capture various types of layout elements is enhanced. This combination of visual prompts and textual prompts helps the model achieve a deeper understanding of webpage layouts. 
\subsubsection{Chain-of-Thought}\label{se2}
In this study, we propose a multi-turn dialogue generation framework based on the CoT strategy. CoT is a step-by-step reasoning approach designed to guide the model in solving complex problems incrementally, aiming to enhance the logical coherence and accuracy of generated answers. Specifically, we generated annotated data from approximately 3,000 high-quality webpage images, using a combination of automated tools and manual curation to annotate clickable elements in the webpage images, such as buttons and links. As shown in Fig. \ref{data_sample}(c), one of the clickable sections of the webpage image is highlighted with a bounding box.
Subsequently, based on the annotations of red bounding boxes, we used the GPT-4o model to generate two rounds of progressive dialogue. The first round of dialogue focuses on the content within the red bounding boxes, generating descriptive QA pairs related to that content. The second round of dialogue further guides the model to predict interactive behaviors, such as inferring the potential content of a page that might appear after clicking a button or link within a given red bounding box, based on the provided prompt.\footnote{More details of visual prompts in the appendix.}\par
Experimental results demonstrate that, with only a small amount of CoT data, the model significantly outperforms baseline models in tasks involving content prediction and interaction behavior inference for webpage images.

\subsection{Datasets}
To support the strategies for visual cue engineering outlined in \ref{se1} and \ref{se2}, we collected data from multiple sources, covering document content, webpage screenshots, OCR text annotations, layout information, question-answer dialogues, and diverse scenarios for reasoning tasks.\footnote{More details of our tranining data in the appendix.}

\begin{table*}[ht]
\renewcommand{\arraystretch}{1.1}
\centering
\caption{The comparison of the RWKV-UI model with open-source general MMLM models and GUI Agent models on VisualWebBench.}
\label{visualbench}
\begin{tabular}{lccccccccc}
\hline
\multirow{2}{*}{\textbf{Model}} &
  \multirow{2}{*}{\textbf{Params}} &
  \multicolumn{3}{c}{\textbf{Website}} &
  \multicolumn{2}{c}{\textbf{Element}} &
  \multicolumn{2}{c}{\textbf{Action}} &
  \multirow{2}{*}{\textbf{Average}} \\ \cline{3-9}
                      &               & Caption       & WebQA         & HeadOCR       & OCR           & Ground        & Prediction    & Ground        &               \\ \hline
\multicolumn{10}{c}{\textbf{Open Sourse MLLMs}}                                                                                                                       \\ \hline
BLIP-2                & 8.2B          & 11.0          & 5.2           & 20.6          & 2.6           & 15.5          & 14.9          & 8.7           & 11.2          \\
mPLUG-Owl2            & 8.2B          & 12.7          & 19.9          & 51.6          & 7.2           & 11.9          & 23.1          & 3.9           & 18.6          \\
Qwen-VL               & 7B            & 21.8          & 32.2          & 48.4          & 13.4          & 14.0          & 26.7          & 10.7          & 23.9          \\
CogVLM                & 17B           & 16.6          & 30.6          & \underline{65.9}    & 10.0          & 17.7          & 11.7          & \underline{23.3}    & 25.1          \\
VILA-13B              & 13B           & 12.7          & 28.8          & \textbf{67.9} & 12.6          & 16.5          & 36.3          & 16.5          & 27.3          \\
DeepSeek-VL-7B        & 7B            & 18.1          & 30.0          & 63.4          & 18.1          & 16.2          & \underline{35.2}    & 15.5          & 28.1          \\
CogAgent-Chat         & 17B           & 16.3          & \textbf{53.3} & 20.2          & 32.4          & \textbf{41.6} & 13.5          & \underline{23.3}    & 28.7          \\
\rowcolor{gray!30}
RWKV-UI-1.6B & 1.6B & 20.4          & 20.4          & 59.9          & 25.9          & \underline{39.5}    & 32.7          & \textbf{34.9} & \underline{33.4}    \\
LLaVA-1.6-7B          & 7B            & \textbf{27.0} & 39.8          & 57.3          & \underline{54.8}    & 31.7          & 30.6          & 10.7          & \textbf{36.0} \\
LLaVA-1.6-13B         & 13B           & \underline{26.5}    & \underline{44.5}    & 52.8          & \textbf{56.1} & 31.7          & \textbf{48.4} & 15.5          & 36.0          \\ \hline
\multicolumn{10}{l}{* \textbf{bold} indicates the highest score, \underline{underlined} indicates the second-highest score.}   
\end{tabular}
\end{table*}

\subsubsection{Screen UI-Related Data}
Screen UI-related data focuses on the layout and content of webpages, leveraging OCR technology to extract text and generate visual prompts that help the model understand the semantics and functions of screen elements. We used Websight\cite{Websight}, Webui-7kbal\cite{Webui-7k}, and Web2Code\cite{Web2Code} datasets, which provide website screenshots, OCR annotations, and interactive UI element screenshots with HTML labels. Webpage layout detection data annotates layout elements (e.g., buttons, titles, images) to help the model understand spatial distribution and hierarchical structure, primarily sourced from the Webui-7kbal and Websight datasets. Web reasoning data simulate user interactions through multiround dialogues to support complex semantic reasoning tasks, using the 1k website screenshots shared by Silatus\footnote{https://huggingface.co/silatus} and the pix2code subset of Web2Code, which are well-suited for interaction element reasoning. Web QA data comes from ScreenQA\cite{Screenqa} and Web2Code, with ScreenQA focusing on QA tasks for mobile page images and Web2Code specializing in QA for PC webpage content.\par
\subsubsection{Document Content and Layout-Related Data}
These data focus on analyzing the structure and content of documents, similar to webpage layout detection but with more emphasis on hierarchical structures and semantic reasoning. Document layout detection data annotates structured content (e.g., titles, paragraphs, tables) to help the model understand document hierarchy, using datasets like PDF-Eng-Image-50k\footnote{https://huggingface.co/datasets/pixparse/pdfa-eng-wds} and IDL-Image-100k datasets\cite{IDL}. Document QA data includes DocLocal4k\footnote{https://huggingface.co/datasets/mPLUG/DocLocal4K}, DocReason25k\footnote{https://huggingface.co/datasets/mPLUG/DocReason25K}, and CORD-V2\cite{cordv2} (for receipt-like documents), covering tasks such as content localization, semantic reasoning, and fine-grained bounding box annotations.\par
\subsubsection{Other Data} General-purpose data focus on tasks less related to webpages or documents, primarily for general QA training. We introduced the Bunny\cite{Bunny} and LLaVAR-Instruct-16k\cite{zhang2023llavar} datasets.\par

\section{Experiments}
In this chapter, we evaluate the performance of our RWKV-UI model on various web-related tasks. First, in \ref{setup}, we introduce the experimental setup, including datasets, model configurations, and the training process. Next, in \ref{results}, we present the main results, using VisualWebBench as the evaluation benchmark. Finally, in \ref{ablation} we conduct ablation experiments to analyze the impact of different components on the model's performance.
\subsection{Experiments Setup}\label{setup}
Our experiments are divided into three training stages: pretraining, domain pretrain, and fine-tuning. Inspired by LLaVA-NeXT\cite{llavanext}, we further pretrain the model on domain-specific data related to web user interfaces and documents. This stage aims to reinforce the model's prior knowledge in UI and document understanding, enabling it to grasp domain-specific vocabulary, structures, and semantics more deeply, thereby improving its performance on downstream tasks in this domain. During both the domain-specific pretraining and fine-tuning stages, we froze the three visual encoders and fine-tuned only the language model.\par

\subsection{Benchmark}
We employ VisualWebBench\cite{visualwebbench} to evaluate the performance of our model. VisualWebBench is a multimodal benchmark that assesses models on webpage understanding tasks using seven key metrics\footnote{More details about the metrics of this benchmark can be found in the appendix.}: Heading OCR, Captioning, WebQA, Action Grounding, Element OCR, Element Grounding, and Action Prediction. The test set comprises instances from 139 real websites covering 87 subdomains.\par

\subsection{Main Results}\label{results}
The test results of the open-source models and RWKV-UI are presented in Tab. \ref{visualbench}: Our model achieved an impressive average score of 33.4, demonstrating exceptional performance with only 1.6B parameters. It outperformed all other models of similar size and rivaled and even surpassed the performance of larger models (e.g., 7B and 13B) on certain metrics. Notably, in the Action Grounding task, our model achieved 34.9, the highest score among all open-source multimodal models with fewer than 17B parameters. This outstanding performance is primarily attributed to the synergistic integration of our proposed COT and layout detection strategies, which enhance the model’s reasoning capability and improve the alignment between visual elements and semantic structures. Furthermore, in the Element Grounding task, our model achieved a score of 39.5, the second-highest overall. This performance highlights the model’s enhanced ability to capture and recognize elements, with visual prompts and high-resolution input playing a key role. Our model's average score of 33.4 also ranks as the second-highest overall, surpassed only by the LLaVA-1.6-13B.\par
This result demonstrates that our strategy enables the RWKV-UI 1.6B model to achieve highly competitive performance in webpage understanding tasks, particularly in UI action prediction, element grounding, and page parsing, despite its smaller parameter size.

\subsection{Ablation Studies}\label{ablation}

\subsubsection{Ablation Study on Resolution Strategy}
\begin{table}[ht]
\renewcommand{\arraystretch}{1.1}
\centering
\caption{The comparison of models with and without High Resolution}
\label{ablation-on-RE}
    \resizebox{1\columnwidth}{!}{
\begin{tabular}{lcccccccc}
\hline
\multirow{2}{*}{\textbf{Method}} &
  \multirow{2}{*}{\textbf{Resolution}} &
  \multicolumn{3}{c}{\textbf{Element}} &
  \multicolumn{2}{c}{\textbf{Action}} &
  \multicolumn{1}{l}{} \\ \cline{3-8} 
                                   &           & \textbf{HeadOCR} & \textbf{OCR} & \textbf{Ground} & \textbf{Prediction} & \textbf{Ground} & \textbf{Average} \\ \hline
Standard Resolution                     & 1024×1024 & 45.1             & 16.5         & 20.1            & 17.8                &\textbf{23.3}            & 22.5             \\
\multicolumn{1}{l}{High Resolution} & 4090×4096 & \textbf{54.0}             & \textbf{23.6}        & \textbf{29.3}           & \textbf{22.4}               & 17.5           & \textbf{26.4}            \\ \hline
\end{tabular}
    }
\end{table}
As illustrated in Fig. \ref{architecture}, the standard resolution strategy relies solely on the global features extracted by the three visual encoders, while the high-resolution strategy combines both global and local features. We compared the performance of models using the standard resolution strategy (1024×1024) and the high-resolution strategy (4096×4096), as shown in Tab. \ref{ablation-on-visualprompt}.
High-resolution input resulted in significant performance improvements across multiple tasks, including Element Grounding (20.1 → 29.3), HeadOCR (45.1 → 54.0), and OCR (16.5 → 23.6), compared to the standard resolution. This improvement arises from the ability of high-resolution input to capture finer visual details, which provides clearer contextual information for element recognition in complex layouts. The overall average score increased from 22.5 to 26.4, highlighting the critical role of high-resolution input in enhancing the model's webpage understanding capabilities.
However, the decline in performance on the Ground task may stem from high-resolution input introducing an overwhelming amount of detail, making it harder for the model to concentrate on webpage action behaviors. Consequently, the incorporation of relevant visual prompts becomes essential to address this limitation and refocus the model.

\begin{table}[ht]
\renewcommand{\arraystretch}{1.1}
\centering
\caption{The comparison of models with and without the visual prompt, Cot and Layout detection strategy}
\label{ablation-on-visualprompt}
    \resizebox{1\columnwidth}{!}{
\begin{tabular}{lcccccc}
\hline
\multirow{2}{*}{\textbf{Method}} & \multicolumn{3}{c}{\textbf{Element}}               & \multicolumn{2}{c}{\textbf{Action}}   & \multicolumn{1}{l}{} \\ \cline{2-7} 
                                 & \textbf{HeadOCR} & \textbf{OCR}  & \textbf{Ground} & \textbf{Prediction} & \textbf{Ground} & \textbf{Average}     \\ \hline
baseline       & 28.5 & 8.9 & 18.2          & 16.4          & 22.3 & 17.0 \\ \hline
+Visual prompt & 54.0             & 23.6        & 29.3           & 22.4               & 17.5           & 26.4  \\
+CoT           & 55.0 & 25.1 & \textbf{43.3} & \textbf{34.5} & 26.2 & 31.9 \\
+Layout detection                & \textbf{59.9}    & \textbf{25.9} & 39.5            & 32.7                & \textbf{34.9}   & \textbf{33.4}        \\ \hline
\end{tabular}
}
\end{table}
\subsubsection{Ablation Study on Visual Prompt}\label{abon-visual-chapter}
To validate the effectiveness of the Visual Prompt and Chain of Thought (CoT) strategies, we conducted rigorous ablation experiments, ensuring consistency in model size, training data, and experimental settings. The Caption and WebQA metrics showed only slight changes, indicating limited impact from these strategies, and were therefore excluded from further analysis. As shown in Tab. \ref{ablation-on-visualprompt}: The introduction of the Visual Prompt significantly improved the model’s ability on HeadOCR(28.5 → 54), Element Ground(18.2 → 29.3) and OCR(8.9 → 23.6), effectively guiding it to understand the positioning and functionality of elements in webpage screenshots.\par
\subsubsection{Ablation Study on CoT Strategy}
With the addition of the CoT strategy, the model first develops a comprehensive understanding of target interactive elements before reasoning, simulating a human-like learning process. This strategy greatly enhances the model’s performance in the Element Ground(29.3 → 43.3) and Prediction(22.4 → 34.5) metrics.\par
\subsubsection{Ablation Study on Layout Detection}
The Layout Detection shows a particularly significant improvement in the Action Ground(26.2 → 34.9) metric. By capturing the spatial relationships of interactive elements and their connections to surrounding regions, as well as providing global page structure information.

\begin{table}[ht]
\renewcommand{\arraystretch}{1.1}
\centering
\caption{The comparison of models with and without Domain pretrain}
\label{ablation-on-DP}
    \resizebox{1\columnwidth}{!}{
\begin{tabular}{lcccccccc}
\hline
\multicolumn{1}{l}{} &
  \multicolumn{3}{c}{\textbf{Websight}} &
  \multicolumn{2}{c}{\textbf{Element}} &
  \multicolumn{2}{c}{\textbf{Action}} &
  \multicolumn{1}{l}{} \\ \cline{2-9} 
\textbf{DP}  & \textbf{Caption} & \textbf{WebQA} & \textbf{HeadOCR} & \textbf{OCR} & \textbf{Ground} & \textbf{Prediction} & \textbf{Ground} & \textbf{average} \\ \hline
\Checkmark   & \textbf{20.4}             & \textbf{20.4}           & \textbf{59.9}             & \textbf{25.9}         & \textbf{39.5}            & \textbf{32.7}                & \textbf{34.9}            & \textbf{33.4}             \\
\XSolidBrush & 20.0          & 18.9           & 42.7             & 9.6          & 33.4            & 27.4                & 29.1            & 25.6             \\ \hline
 
\end{tabular}
    }
\end{table}

\subsubsection{Ablation Study on Domain Pretrain}
As show in Tab. \ref{ablation-on-DP}: In the ablation study of Domain Pretrain(DP), the results demonstrate that domain pretrain plays a crucial role in improving the model's subsequent training process. In specialized domain training, the model is often exposed to a continuous input of new domain-specific knowledge, which increases the complexity of learning and makes training more challenging. Domain pretrain addresses this issue by injecting domain-specific prior knowledge into the model, enabling it to establish an initial understanding of the domain and reducing its reliance on learning everything during subsequent training stages. This prior knowledge allows the model to adapt more efficiently to complex tasks in later training stages, especially in tasks such as OCR and Grounding, where the model demonstrates significantly improved precision in understanding page structures and interactive elements.

\section{Conclusion}
This study presents RWKV-UI, a visual language model specifically designed for high-resolution user interface (UI) understanding. With a three-encoder architecture and partition encoding strategy, combined with visual prompt engineering and the CoT mechanism, the model outperforms existing models of similar scale in both performance and reasoning capabilities. It provides a reliable technical framework for high-resolution UI tasks and holds potential for further applications in multimodal scenarios.

\bibliographystyle{IEEEbib}
\bibliography{main}

\vspace{12pt}

\newpage
\onecolumn
\appendix
\subsection{Examples of visual prompt}
\begin{figure}[ht]
    \centering
    \includegraphics[width=1\linewidth]{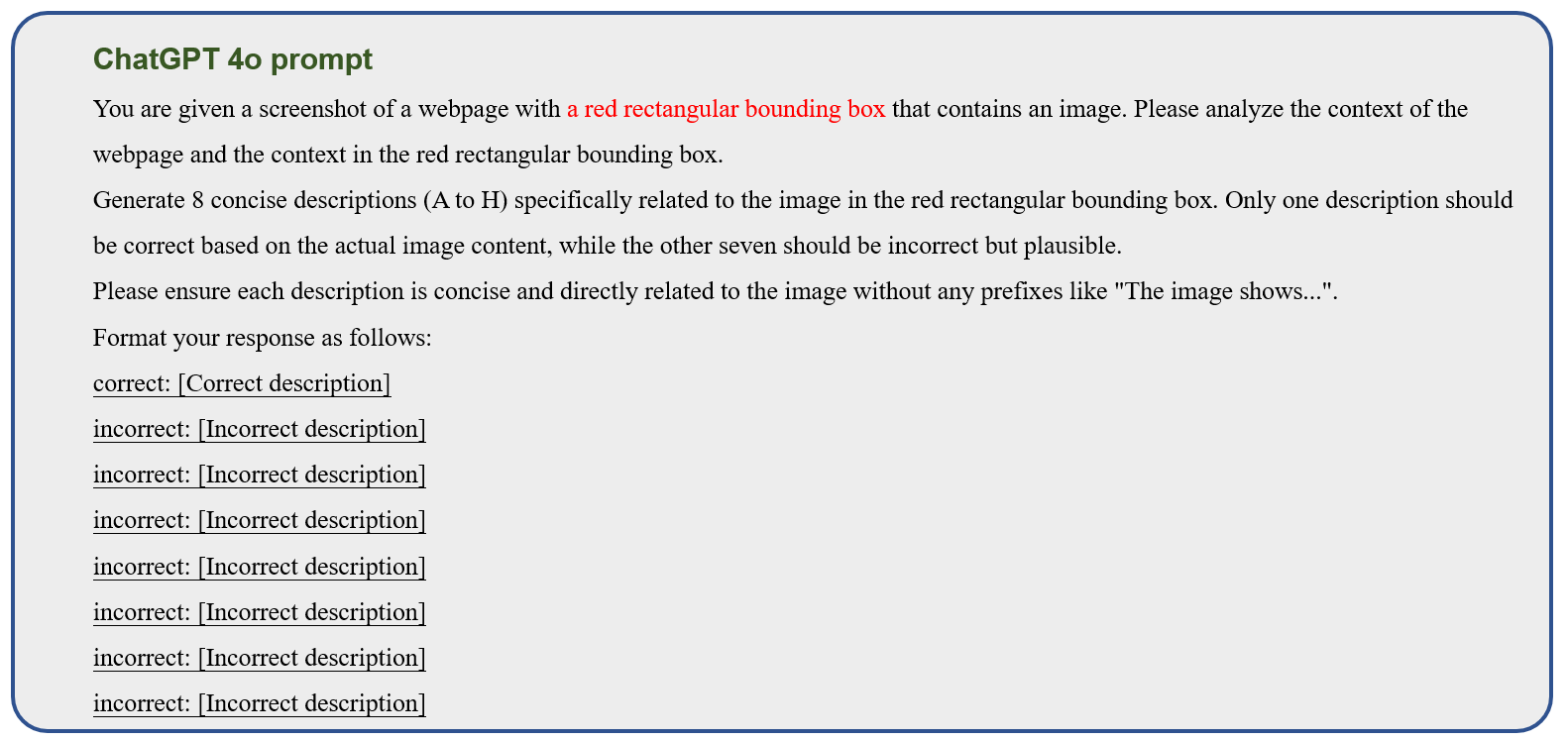}
    \caption{Prompts for calling the GPT API.}
    \label{gptprompt}
\end{figure}

\begin{figure}[ht]
    \centering
    \includegraphics[width=1\linewidth]{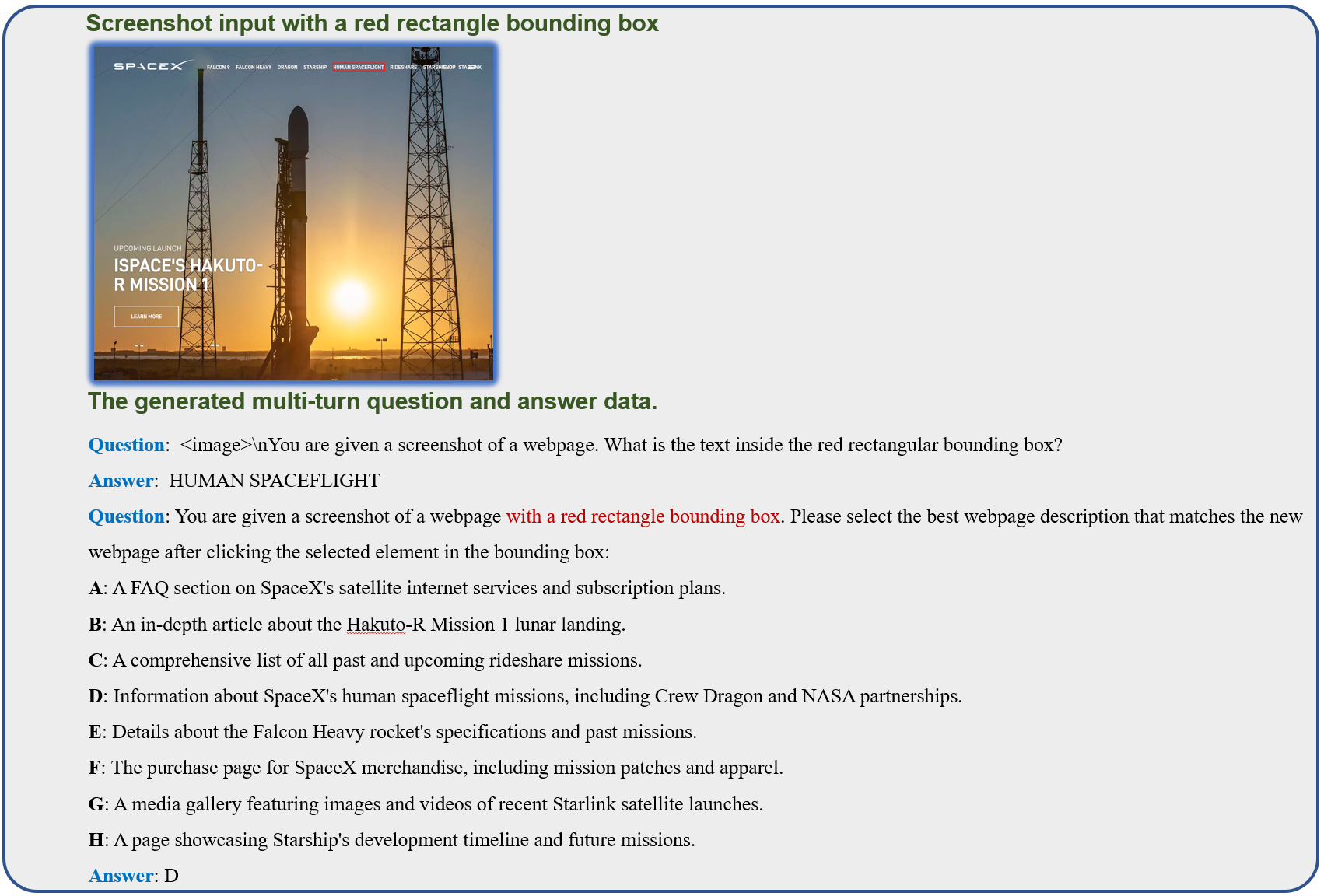}
    \caption{An example of CoT data generated by calling the ChatGPT 4.0 API.}
    \label{cotsample}
\end{figure}

Here, we present examples of the CoT visual prompt data we designed, as shown in Fig. \ref{gptprompt} and \ref{cotsample} We utilized the ChatGPT-4 API to generate corresponding prediction data for webpage behaviors, incorporating both recognition and reasoning chains: First, we let GPT4o identify the text or image content in the web page pictures annotated with red rectangular frames, and conduct a simple OCR and content understanding. Subsequently, based on this content, we let GPT4o predict the web page content after a clicking behavior.

\subsection{Train details}
During the pre-training stage, we used 558k data from LLaVA for pre-training. We conducted the training on four 4090 GPUs with 24G of running memory. A single epoch took approximately 20 minutes. Due to limited memory, we set the minimum batch size to 1 and carried out the training for 140 epochs. It took about 46.5 hours for the pre-training. During the pre-training stage, we froze 24 layers of RWKV. The initial learning rate was 1e-3, and the data precision type was bfloat16.\par
In the domain pre-training stage, we utilized 193k data that we collected and processed, including abundant chart-text and basic web question-and-answer data. We trained for 48 epochs, which took 16 hours. The initial learning rate was 6e-5, and we unfroze 24 layers of RWKV model.\par
Finally, in the fine-tuning training stage, we adopted 555k data containing visual prompts. The learning rate was 6e-5, and the batch size remained 1. We trained for 139 epochs, which took about 46 hours.
\subsection{Datasets}
 The specific data components used in the three training stages are provided in Tab. \ref{training data}.

\begin{table}[ht]
\renewcommand{\arraystretch}{1.1}
\begin{center}
\caption{Training data statistics}
\label{training data}
\begin{tabular}{lllcc}
\hline
\textbf{Training State}                     & \multicolumn{2}{l}{\textbf{Dataset Name}} & \textbf{Num} & \textbf{Sum}                   \\ \hline
Pretrain & \multicolumn{2}{l}{LLaVA Instruct}                & 558k & 58k \\ \hline
\multirow{6}{*}{Domain Pretrain} & \multicolumn{2}{l}{cord-v2}      & 0.8k      & \multirow{6}{*}{193k} \\
         & \multicolumn{2}{l}{DocLocal4k}          & 4.2k &     \\
         & \multicolumn{2}{l}{DocReason25k}        & 25.7k &     \\
         & \multicolumn{2}{l}{PDF Association dataset}                 & 40k  &     \\
         & \multicolumn{2}{l}{Industry Documents Library}                 & 15.5k  &     \\
         & \multicolumn{2}{l}{Websight}            & 107k &     \\ \hline
\multirow{9}{*}{Fine Tuning}             & \multicolumn{2}{l}{web2code}     & 35k       & \multirow{9}{*}{555k} \\
         & \multicolumn{2}{l}{Bunny}                & 140k &     \\
         & \multicolumn{2}{l}{websight}            & 220k &     \\
         & \multicolumn{2}{l}{PDF Association dataset}                 & 9.3k &     \\
         & \multicolumn{2}{l}{Industry Documents Library}                 & 7.6k &     \\
         & \multicolumn{2}{l}{screenqa}            & 33k  &     \\
         & \multicolumn{2}{l}{LLaVAR-Instruct-16k} & 33k  &     \\
         & \multicolumn{2}{l}{webui-7kbal}         & 75k  &     \\
         & \multicolumn{2}{l}{1k-websight}         & 1k   &     \\ \hline
\end{tabular}
\end{center}
\end{table}
\subsection{Results on VisualWebbench}
\begin{figure}[ht]  
 \begin{minipage}{0.48\linewidth}
     \vspace{3pt}  
     \centerline{\includegraphics[width=\textwidth]{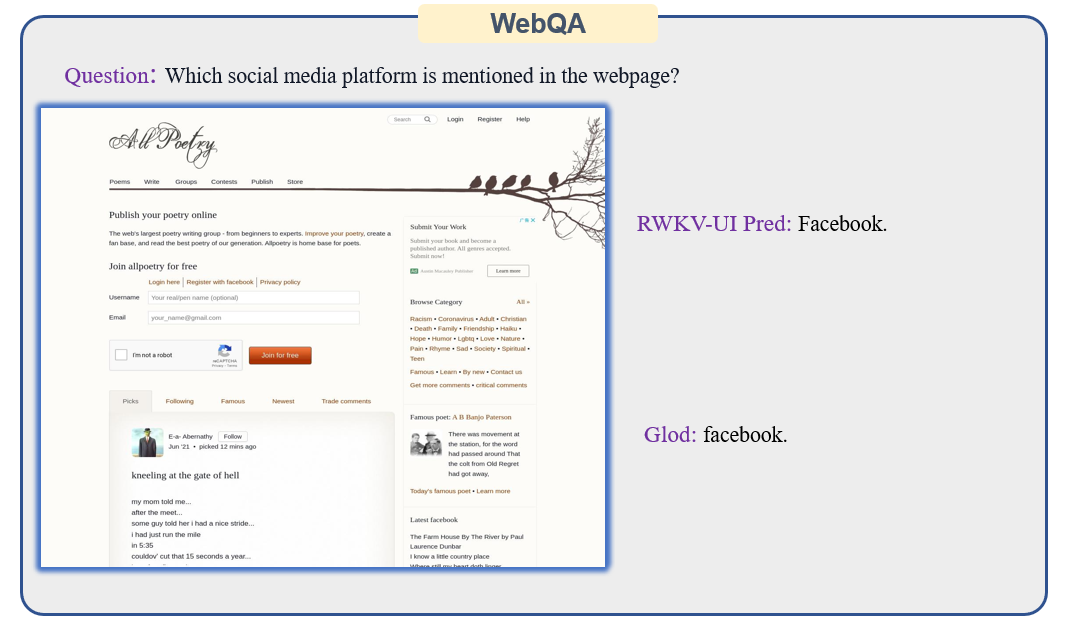}}
     \vspace{3pt}
     \centerline{\includegraphics[width=\textwidth]{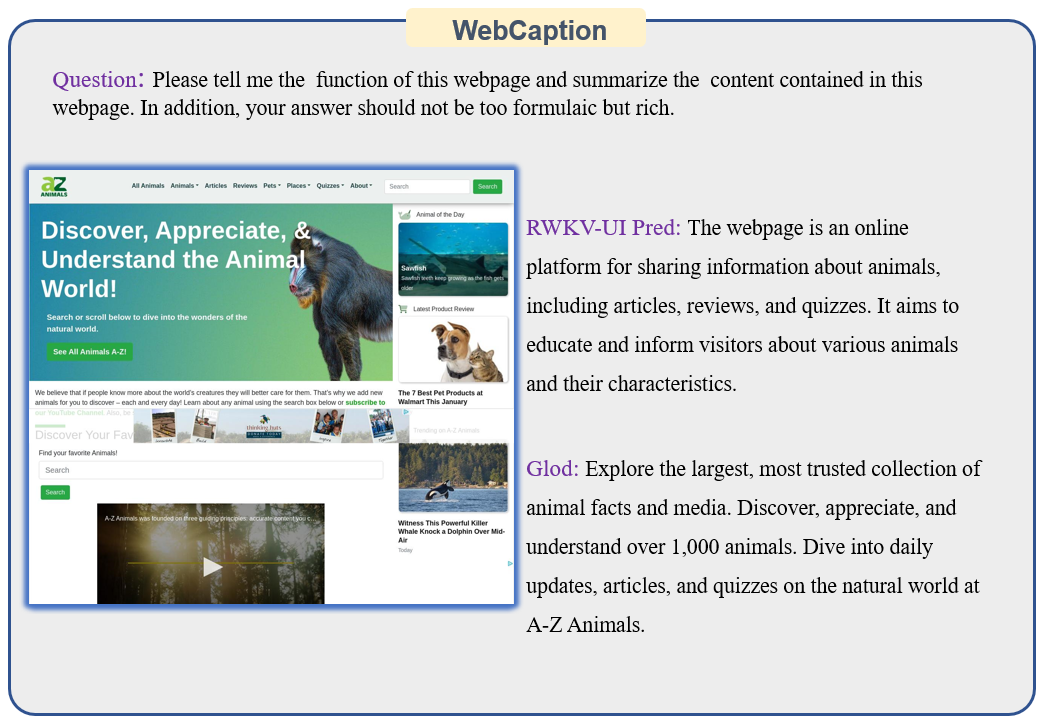}}
     \vspace{3pt}
     \centerline{\includegraphics[width=\textwidth]{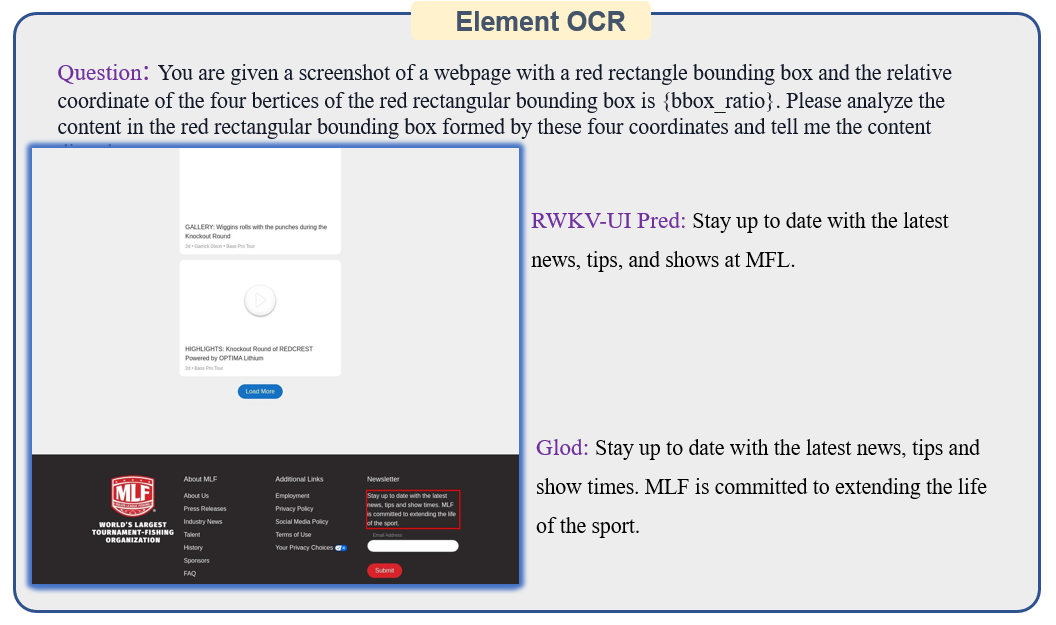}}
     \vspace{3pt}
     \centerline{\includegraphics[width=\textwidth]{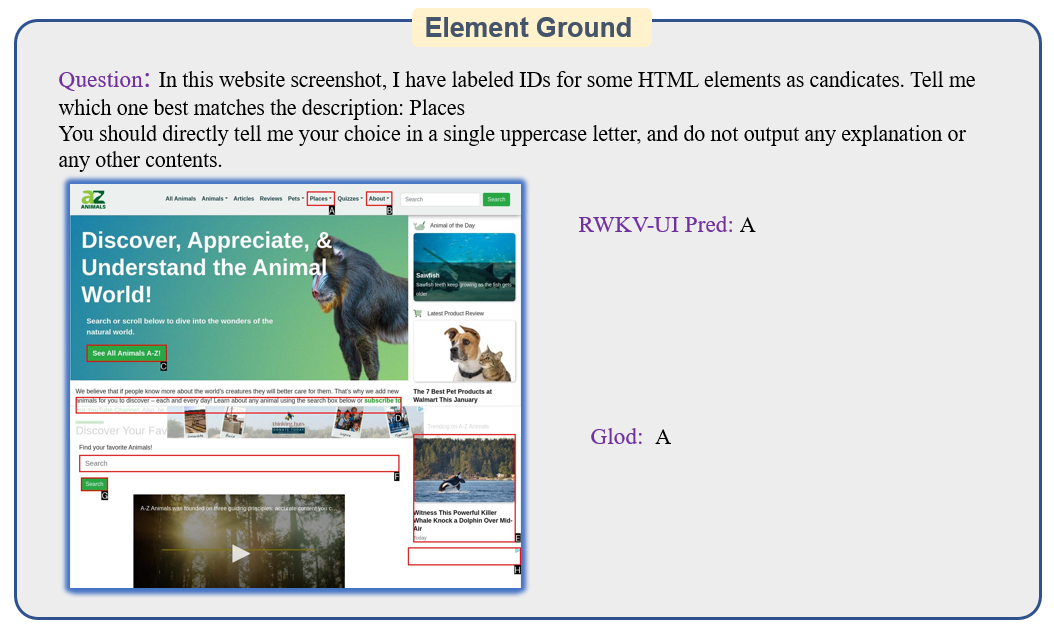}}
     \vspace{3pt}
\end{minipage}
    \begin{minipage}{0.48\linewidth}
     \vspace{3pt}
     \centerline{\includegraphics[width=\textwidth]{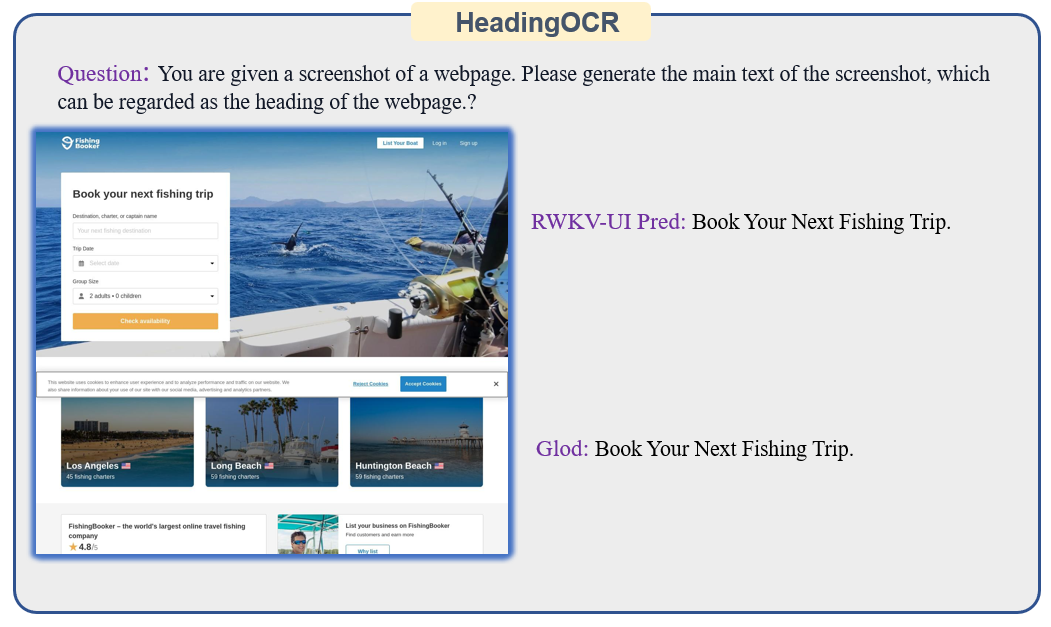}}
     \vspace{3pt}
     \centerline{\includegraphics[width=\textwidth]{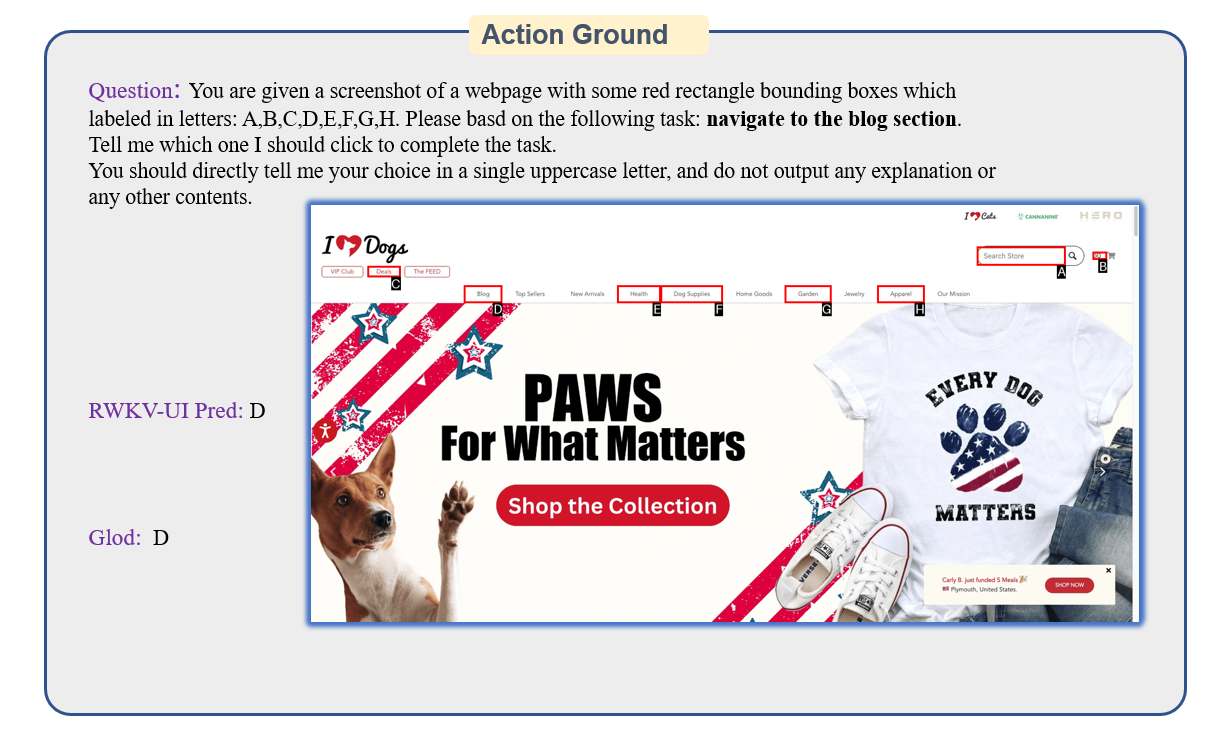}}
     \vspace{3pt}
     \centerline{\includegraphics[width=\textwidth]{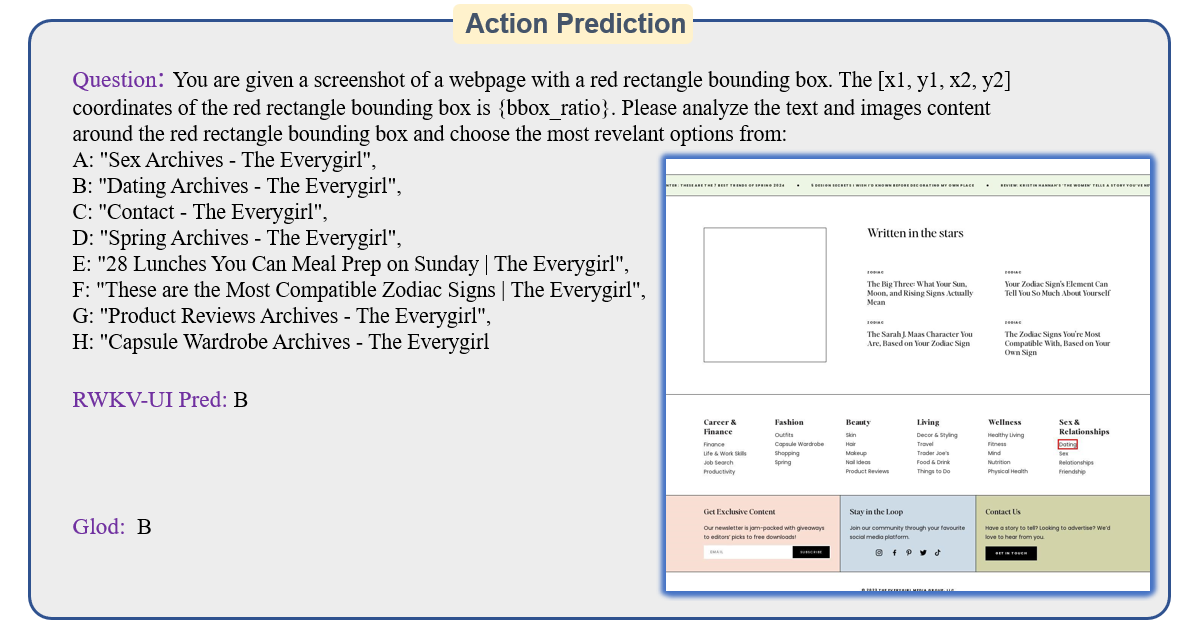}}
 \end{minipage}
	\caption{The results on seven metrics.}
	\label{benchresult}
\end{figure}

Next, we will present the test results of the RWKV - UI model on different metrics of VisualWebbench, as shown in Fig. \ref{benchresult}. The figure shows the test data and results of seven metrics: Heading OCR, Captioning, WebQA, Action Grounding, Element OCR, Element Grounding, and Action Prediction. Among them, the bbox\_ratio in the test examples of the Element OCR and Action Prediction metrics represents the relative coordinates of the red bounding box within the webpage image.

\end{document}